\documentclass[10pt,twocolumn,letterpaper]{article}

\usepackage{wacv}
\usepackage{times}
\usepackage{epsfig}
\usepackage{graphicx}
\usepackage{amsmath}
\usepackage{amssymb}
\usepackage{adjustbox}

\usepackage[accsupp]{axessibility}  

\usepackage {nopageno}
\def\wacvPaperID{485} 

\wacvfinalcopy 

\ifwacvfinal
\fi

\ifwacvfinal
\usepackage[breaklinks=true,bookmarks=false]{hyperref}
\else
\usepackage[pagebackref=true,breaklinks=true,colorlinks,bookmarks=false]{hyperref}
\fi

\begin{document}

\title{Single Image Object Counting and Localizing using Active-Learning}

\author{Inbar Huberman-Spiegelglas
\qquad
Raanan Fattal\\
{\tt\small \{inbar.huberman1, raanan.fattal\}@mail.huji.ac.il}\\
School of Computer Science and Engineering \\ The Hebrew University of Jerusalem, Israel\\

}

\maketitle

\begin{abstract}
The need to count and localize repeating objects in an image arises in different scenarios, such as biological microscopy studies, production-lines inspection, and surveillance recordings analysis. The use of supervised Convolutional Neural Networks (CNNs) achieves accurate object detection when trained over large class-specific datasets. The labeling effort in this approach does not pay-off when the counting is required over few images of a unique object class.

We present a new method for counting and localizing repeating objects in \emph{single-image} scenarios, assuming no pre-trained classifier is available. Our method trains a CNN over a small set of labels carefully collected from the input image in few active-learning iterations. At each iteration, the latent space of the network is analyzed to extract a minimal number of user-queries that strives to both sample the in-class manifold as thoroughly as possible as well as avoid redundant labels.

Compared with existing user-assisted counting methods, our active-learning iterations achieve state-of-the-art performance in terms of counting and localizing accuracy, number of user mouse clicks, and running-time. This evaluation was performed through a large user study over a wide range of image classes with diverse conditions of illumination and occlusions.
\end{abstract}

\section{Introduction}
Localizing and counting repeating objects is useful for various purposes, such as counting cells under the microscope, inspecting products in production-lines, and tracking objects in surveillance cameras. These tasks typically require a considerable amount of repetitive human labor, and in many cases, a high degree of attention in identifying the object of interest despite variations in its appearance or a cluttered background.

Traditional computerized approaches for detecting and counting objects are based on handcrafted visual descriptors which are either class-specific~\cite{Chen2012,Idrees2013} or assume the object appearance is well-resolved~\cite{Huberman2016}. While these methods are efficient and require little or no training, they have a limited ability to cope with the complicated visual variability that real data may contain. 

Convolutional Neural Networks (CNNs) achieve very high accuracy in object recognition~\cite{Girshick2014,Krizhevsky2012} and detection tasks~\cite{Hu18}. Nevertheless, this discriminatory power is achieved at a non-trivial cost of collecting datasets containing large numbers of manually-annotated images for every image class. Such fully-supervised CNNs are used for detecting and counting repeating objects in an image~\cite{Cohen2017,Hsieh2017,Walach16}. This approach is practical only when the networks are used extensively at inference time over the same image class they were trained for. In cases where there is a single image, containing several dozens up to few hundreds of object appearances, a fully-supervised training will require the same amount of manual effort as the counting task itself. Hence, an alternative approach is needed for utilizing CNNs in such scenarios.

In this paper, we describe a new network based method that is applicable for single image object detection and counting scenarios. The method collects a small number of labels from the input image which allow it to cope and specialize over novel objects of interest. 




The method achieves accurate detection coordinates and counting estimate despite the small number of training examples it uses by targeting highly informative training labels. The latter are obtained using an iterative active-learning scheme consisting of a query sampling step, a label collection step that minimizes the user effort, and a network re-training step that accounts for the newly labeled examples. 


In order to minimize the number of labels and user effort, at every iteration, our query sampling strategy follows two basic principles: it targets regions where previously collected labels do not generalize well, as well as avoids queries that are expected to receive a coupled classification. The key novel aspect of our construction is in performing these steps and analyses over the activations in the network being trained. These neural responses provide a high-dimensional descriptors, known as visually-meaningful cues~\cite{Johnson16,Kupyn2018,wang2018}. Existing object counting approaches either rely on the scalar output of the classifier, or do not follow any sampling strategy.

Specifically, by clustering these activations we obtain a transitivity relation in the image which we use for detecting regions (clusters) lacking a label as well as avoid multiple queries over closely-related regions (same cluster). To further minimize the user labeling effort, we present the query windows along with a tentative classification, allowing the user to mark only the misclassified cases.


We report the results of a user-study that demonstrates the proposed method's ability to achieve state-of-the-art accuracy on real and synthetic benchmark images. The images used in this study are diverse, span many object classes, and contain challenging variations in shape, illumination, and occlusion relations. Finally, we show that the network can be trained on one image and used for counting in other images of the same type and source.
\section{Related Work}
The problem of object counting and localization in images with repeating objects received a considerable attention, and a variety of visual models were suggested. Earlier approaches tackled specific image classes, such as crowd, cars, and cells, and used domain-specific features to perform the counting. Some examples of crowd counting approaches include the use of edge orientations and blobs size histograms in~\cite{Kong06}, texture descriptors with frequency-domain analysis in~\cite{Idrees2013}, and segmentation-based feature extraction in~\cite{Chan08}. Object localization in this context is described in~\cite{Chen2012,Dong07,quteprints09}.

A more general approach, developed by Lempitsky and Zisserman~\cite{Lempitsky2010}, estimates the objects density map in a class non-specific manner. This method learns a density regression over a number of general features in a supervised fashion. To alleviate the need for large annotated training sets, Arteta \etal~\cite{Arteta14} proposed a system for counting and localizing objects where the user provides dot annotations in an interactive session displaying the inferred counting. While this method shows very good performance over diverse classes, the use of density estimation undermines the localization accuracy. Huberman and Fattal~\cite{Huberman2016} also localize and count objects by exploiting the repetitions in the image to automatically fit a deformable-part model which is fed into a user-guided classifier. This method is challenged in cases where the repeating patches in the image do not necessarily correspond to the repeating object. The work of von Borstel \etal.~\cite{Borstel16} also counts object repetitions using a fairly small number of region-level count labels. The method models the object density using a Gaussian process that applies non-linear kernels on various image features. In contrast to our method, this approach does not provide object localization.

More recent works employ the detection power of CNNs for object counting. Ma \etal.~\cite{Ma15} extend the density map approach of~\cite{Lempitsky2010} to cope with small-instance scenes. In this work integer programming over parts of the density map is used to recover the object locations. Walach and Wolf~\cite{Walach16} improve the counting accuracy using network boosting and selective sampling which reduce the effect of high error samples by temporally muting them. Kang \etal~\cite{Kang17} compare crowd density maps, generated by several density estimation methods, and their performance on visual tasks including counting, detection and tracking. They also show how the detection accuracy can be improved by computing a full-resolution density map, as opposed to the reduced-resolution maps used by some methods. Unlike these fully supervised methods, we aim for a weakly-supervised solution which enables object localization and counting given a single-image.

Segu{\'{\i}} \etal~\cite{SeguiPV15}, train a counting network and use its deep features to derive a confidence map which they use for detecting object occurrences. Dijkstra \etal~\cite{Dijkstra18} train a network that outputs several channels which are fed into a voting mechanism that estimates the object centers robustly. Xie \etal~\cite{Xie16} train a CNN to output a density map and use it for cell localizing. Our network is trained to predict objects occurrences using a sparse set of training coordinates. 




Loy \etal~\cite{Loy13} try to minimize the user input in the specific context of crowd videos by restricting the annotations to the most informative frames. Cao \etal, ~\cite{Cao2017} also attempt to reduce this manual effort by describing an annotation scheme that alleviates the need to explicitly mark bounding boxes over images videos of vehicles. Finally, Lu \etal~\cite{Lu2018} reduce the amount training examples needed by using a pre-trained general-purpose matching network, which can be further refined for a specific object class given a smaller set of examples. We minimize the user effort by requesting the correction of a small set of highly-informative image windows. The latter are presented to the user along with their tentative classifications produced by our network during the iterative active-learning procedure.

\begin{figure}[t]
\centering
\includegraphics[width=3.5in]{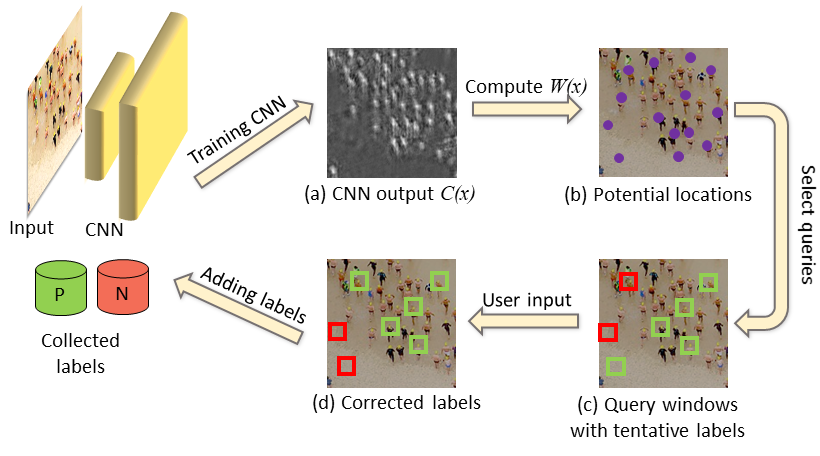}
\caption{Method Overview. At each iteration the network is given with the input image (people in a beach in this example) along with the labels collected so far, and it is trained to predict this sparse set of labels in its classification map $C(\textbf{x})$, as shown in (a). A non-maximum suppression is applied over the positive values of this map to obtain a list of potential object appearances $\mathcal{W}$, as shown in (b) by the purple dots. By analyzing the latent representation of these locations we select a small number of informative query windows. We use the network being trained in order to predict tentative labels to present to the user, as shown in (c). Next, the user corrections are obtained, shown in (d), and are added to the labels training sets, $\mathcal{P}$ and $\mathcal{N}$. The network is retrained on these updated sets and the process repeats itself.}
\label{fig:flow}
\end{figure}






\newcommand {\raanan}[1]{\bf{Raanan: #1}\normalfont}

\newcommand {\posl}{$\mathcal{P}\: $}
\newcommand {\negl}{$\mathcal{N} $}
\newcommand {\wwp}{$\mathcal{W}^\mathcal{P} $}
\newcommand {\wwn}{$\mathcal{W}^\mathcal{N} $}

\newcommand {\subs}[1]{\bf{#1.}\normalfont}
\section{New Method}
\label{sec:method}
An image containing a repeating object, by its definition, provides a small yet highly-relevant dataset for characterizing the object's appearance. When the number of repetitions is sufficiently large, compared to the complexity of its visual variations, generalization is made possible. Our method operates according to this observation and relies solely on the input image as its training data. As we describe below, the user is required to mark the object of interest in the image.



We use a fully-convolutional neural network to predict the likeliness of an object occurrence at every image pixel. The network is trained in conjunction with collecting its training labels via an iterative active-learning procedure consisting of the following steps. We use the partially-trained network to extract a list of potential object occurrences based on local maxima in its classification map. To minimize the user effort in validating all these suspected locations, we extract highly-informative subset for hers/his input. More specifically, our sampling strategy discards similar regions as well as regions that appear close to ones already labeled by analyzing the latent representation of the image in the network.


This small set of highly-informative queries are presented to the user along with tentative labels using color-coded frames around the suspected pixels. Thus, the user is not required to label all the windows but only click over the ones requiring a label correction. Given these new labels, additional training steps are applied until the network conforms to all the labels collected. As we describe below, the tentative labels are derived from the activations of the network being trained.

Figure~\ref{fig:flow} illustrates this
cycle, which repeats itself until it is terminated by the user. We proceed by describing each of these steps in detail, and start by the initialization of this iterative procedure.

\subsection{Initialization} 
\label{sec:init}
As an input, we assume a single image containing multiple repetitions of the object of interest. In order to specify the repeating object, the user is asked to mark a bounding-window, $B$, and the image is then rescaled such that $B$ becomes 21-by-21 pixels. In case of several repeating objects, the user is required to mark a window around each object type and we rescale the input according to the largest marked window. Being a fully-convolutional network, our classifier can be applied over any image size. We denote the input image by $I(\textbf{x})$, where $\textbf{x}\!:=\!(x,y)$ are pixel coordinates. 



As noted above, the procedure we apply is iterative and assumes a partially-trained classifying network is available at each iteration. We initialize the network by training it over sets of positive \posl and negative \negl pixel labels derived from the Normalized Cross-Correlation (NCC) function between the input image $I$ and the user-provided window $B$. Due to the limited ability of the correlation function to achieve a reliable detection, we use it very conservatively and set
\begin{equation}
\label{eq:init_labels}
\begin{split}
\mathcal{P}&=\{ \textbf{x} : MaxSup\big(NCC(I,B)\big)( \textbf{x}) >0.85 \} \\ \mathcal{N}&=\{ \textbf{x} : NCC(I,B)( \textbf{x}) < 0 \},
\end{split}
\end{equation}
where $MaxSup$ denotes a non-maximum suppression operator computed within windows of 11-by-11 pixels (half the bounding-window size). This suppression window size is chosen in order to prevent extracting multiple detections of the same object appearance, yet allows the identification of partially occluded instances. The threshold of 0.85 was found optimal in a hyper-parameters study we report in the Appendix Section in the Supplemental Material accompanying this submission. In the following active-learning iterations, more labels will be collected from the user and added to these initial sets \posl and \negl.

While a single window is typically sufficient for this initialization, experiments show that at least $10$ positive coordinates in \posl are required (for each repeating object type) for the classifier to properly initialize. Hence, if necessary, the user is asked to provide additional bounding-windows until $10$ positive examples are obtained from their correlation function. Finally, due to the immediate correspondence between the suspected pixel coordinates and the windows around them, we refer to both interchangeably.

At every iteration, including the initialization step, we train the network to output a classification map $C(\textbf{x})$ with positive values over the pixels in \posl and negative over \negl, by minimizing the following label loss
\begin{equation}
\label{eq:label_loss}
\begin{split}
L_{label} = \sum_{\textbf{x}\in\mathcal{P}} \dfrac{(C(\textbf{x}) - 1)^2}{|\mathcal{P}|} + \sum_{\textbf{x}\in\mathcal{N}} \dfrac{(C(\textbf{x})+ 1)^2}{|\mathcal{N}|}.
\end{split}
\end{equation}
The minimization steps are applied until $C(\textbf{x})\geq 0.95$ over $\textbf{x} \in \mathcal{P}$ and $C(\textbf{x}) \leq -0.95$ over $\textbf{x} \in \mathcal{N}$. 
This loss does not pose any requirement over $C$ at the rest of the image pixels. 

\begin{figure*}[t]
\centering
\includegraphics[width=5in]{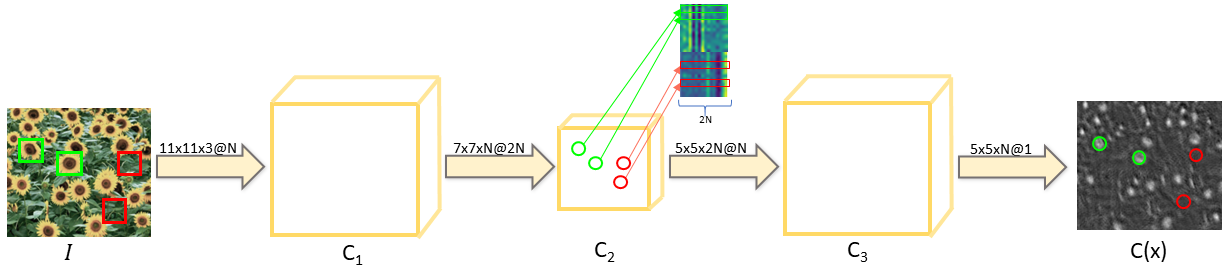}
\caption{The classification network consists of three hidden layers where the latent representation vectors are taken from its deepest layer, $C_2$. Left to right are: the input image with two query windows from \wwp and two from \wwn shown with their appropriate colored frames. The latent vectors of the query windows are shown next to the deepest layer, where the restriction to the positive and negatives sub-spaces is apparent. At the right we see the current classification map $C(\textbf{x})$, where the labels obtained are circled. Note that in this example no user correction is needed.} 
\label{fig:network}
\end{figure*}

\textbf{Network Architecture and Training. }\normalfont In our implementation, we use an ADAM optimizer with a learning rate of $10^{-3}$. As we report in Section~\ref{sec:results}, the use of MSE loss in Eq.~\ref{eq:label_loss} provides better performance than a cross-entropy loss. An example classification map $C(\textbf{x})$ is shown in Figure~\ref{fig:network} along with its sparse sets of training points.

The classifying network we use consists of a fully-convolutional neural network with two \emph{encoding} layers of convolution, bias and ReLU operators. A 2-by-2 max-pooling operation is applied only at the second layer. These layers are followed by another two \emph{decoding} layers (with a 2-by-2 unpooing step in the first decoding layer) that produce the network's output classification map, $C(\textbf{x})$, to have the same resolution as the input image. The ReLU operator is omitted from the last layer to permit negative classification values. The max-pooling switches are kept and used by the unpooling operator. Moreover, the number of filters (channels) in the network grows and then shrinks and is given by $c_{in}\rightarrow N\rightarrow 2N \rightarrow N \rightarrow 1$, where $c_{in}$ is the number of input image color channels (typically, 1 or 3). Figure~\ref{fig:network} summarizes the network architecture used.




Since we want to minimize the number of training labels, we carefully set the number of filters $N$ based on the visual complexity of the image.
We estimate the latter by defining an auto-encoder (AE) with an architecture similar to our classification network, and search for the value of $N$ that achieves a sufficient degree of image reproduction. 

More specifically, we add another 2-by-2 max-pooling step at the end of the first layer of our classifying network, and its corresponding unpooling step in the last decoding hidden layer. Since the AE reconstructs the input image, the number of filters in its output layer is set to $c_{in}$. We start this search with $N=8$ and train the AE to minimize the reconstruction loss, $E_{rec}=\|AE(I)-I\|_2$. In case that $E_{rec} > 10^{-2}$, means that the power of the network is not enough for reconstruction, we increase $N$ by 8 and repeat this process.



\subsection{Sampling User Queries} 
\label{sec:selection}

Given the input image and the partially trained network, at each iteration of the active-learning process we sample a new set of query windows and present them to the user for correction. Unlike the standard approach, used in~\cite{Huberman2016}, we do not carry out this query selection based on the classifier's output, but operate in the activation space of the network.


Since the set of positive labels \posl was initialized by a fairly-conservative NCC threshold, we concentrate the query extraction on collecting additional positive labels. Thus, the queries are selected from a list of pixel locations, $\mathcal{W}$, where the network's classification is positive (suspected object occurrences), i.e., $\mathcal{W}=MaxSup(C(\textbf{x}))\geq0$, where again a suppression window size of 11-by-11 pixels is applied. Clearly any previously labeled coordinate is omitted from $\mathcal{W}$. Figure~\ref{fig:flow} shows an example of such potential locations.

Finding the set of query windows from $\mathcal{W}$ that will be most informative for further network training is the key for minimizing the user effort involved in this process. Hence, to efficiently sample the in-class manifold, we: (i) target the search toward windows that appear to the network least similar to ones that were already labeled, as well as (ii) avoid querying, at the same round of user feedback, multiple windows which the network finds similar and are likely to receive a dependent classification. 

Recent studies show that the latent representation of classification CNNs can be used to provide visually meaningful metric spaces~\cite{Johnson16,Zhang18}. In addition, Caron \etal.~\cite{Caron18} show that this representation is an effective space for proximity measure even at the early stages of the training, including the initialization. Consequently, during training we carry out both these proximity considerations over the deepest encoding layer, $C_2$, which correspond to a high-dimensional feature extraction. We proceed with detailed description of these query sampling steps.


\subs{Avoid Redundant Queries} Being a continuous function, coordinates with close latent vectors are likely to obtain a similar classification by the subsequent decoding layers. Hence, we cluster the pixel coordinates in $\mathcal{W}$, by computing a $k$-means clustering, $\{\Theta_i\}_{i=1}^k$, using $L_2$ norm over their corresponding latent vectors. Note that the spatial resolution of $C_2$ differs from the input resolution by one 2-by-2 pooling step. However, this does not undermine the one-to-one correspondence between the coordinates in $\mathcal{W}$ and the latent vectors due to the non-maximum suppression that ensures a sufficient spacing inside $\mathcal{W}$. We use the resulting clusters in order to avoid redundant queries by limiting selection from $\mathcal{W}$ to no more than one coordinate from the same cluster.

\subs{Obtaining Informative Queries} Moreover, we use this clustering to avoid querying windows which are similar to ones already labeled, and by that explore novel and poorly-generalized regions in the in-class manifold. This is done by computing the $L_2$ distance $d_w$ between the latent vectors of every potential coordinate $w \in\mathcal{W}$ and its closest labeled window in $\mathcal{P} \cup \mathcal{N}$. Then, at each cluster $\Theta_i$ we extract the window $q_i$ that is farthest from any user-provided label window, by picking $q_i = \text{argmax}_{w \in\Theta_i} d_w$.



\subs{Targeting the Separation Margin} As aimed, this choice results in a single, poorly-generalized, query from each cluster. Recall also that all these queries are derived from $\mathcal{W}$ and receive a positive classification from the network, i.e., $C(\textbf{x})>0, \forall \textbf{x} \in \mathcal{W}$. As noted above, this allows us to target and enrich the set of positive labels \posl. However, during the training process the network classification is unreliable and some of the coordinates in $\mathcal{W}$ are likely to be false-positives.

The user-provided labels for these queries is meant to resolve these errors, nevertheless we aim for a weak user supervision. Therefore, we follow the strategy in~\cite{Scheffer01} and better target the query selection towards the separation margin between the true- and false-positive in $\mathcal{W}$.


We use this space to split $\mathcal{W}$ into \wwp and \wwn based on their proximity to labeled pixels in \posl or \negl respectively, i.e., a point in $\mathcal{W}$ will be inserted to \wwp if it has a closer point (in the latent space) in \posl than in \negl, and vice versa.

\begin{figure}[t]
\centering
\includegraphics[width=2.8in]{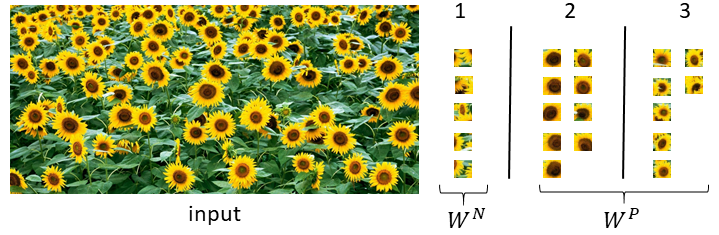}
\caption{This example shows three clusters, two from \wwp and one from \wwn, computed over the activations of the image in the network. Each cluster consists of similar visual windows.}
\label{fig:kmeans}
\end{figure}

Finally, we apply the process described above over \wwp and \wwn separately, i.e., we compute $k$-means clustering over each set, then extract a single poorly-generalized window from each cluster, and present the user with the five top scoring ones (based on their $d_w$) from \wwp and another five from \wwn. In order to obtain five meaningful positive and negative queries, we extract twice this number of clusters, i.e., compute the $k$-means with $k=10$ over each set. Figures~\ref{fig:kmeans} shows example clusters extracted in this process. These clusters were computed right after the initialization step, showing the visual-meaningful metric space of the activations at the very beginning of the training process. Compared to a random query selection (from \wwp and \wwn), our clustering-based query retrieval reduces the average error percentage by 12.9\% as shown in the ablation study we report in the Appendix Section in the Supplemental Material accompanying this submission.

\subs{Sub-Space Separation in Latent Space} The query extraction procedure described above assumes the $L_2$ distance in latent space provides a reliable visual proximity measure. We further conform the latent space representation with this metric by introducing a disjoint sub-space constraint.

Specifically, we add an additional loss term, on top of the label loss in Eq.~\ref{eq:label_loss}, that restricts the latent representation of the positive and negative windows to be mapped to disjoint sub-spaces, by
\begin{equation}
\label{eq:subspace_loss}
\begin{split}
L_{sub} = & \sum_{x\in\mathcal{P}} \|C_2(\textbf{x})[1\!:\!m]\|_2^2 +  \sum_{x\in\mathcal{N}}  \big\|C_2(\textbf{x})\big[(m+1)\!:\!2N\big]\big\|_2^2, \\ & \text{such that} \: \|C_2(\textbf{x})\|_2=1 \: \text{at every point} \: \textbf{x},
\end{split}
\end{equation}
where $C_{2}$ denotes the deepest latent activation vectors and $[\cdot]$ denotes their channels. This loss attempts to map the negative windows through the first $m$ channels of the latent vectors (by minimizing the rest), and the positive windows through the last $2N-m$ channels. The unit-norm constraint ensures no trivial solution is reached, and is carried out by a simple normalization layer coded into the network graph. Note that the choice of which channels are associated with which group (positive or negative) is arbitrary and has no effect over this separation.

We evaluate the effect of $m$ in the ablation study and show it has a little influence. We opted for $m=N$ in our implementation. Figure~\ref{fig:network} shows example activation vectors revealing the restrictions produced by this sub-space term. We also report in the ablation study the effect of this loss term addition, which contributes a 13.7\% reduction in average percentage counting error.

\subsection{User Input} 

Given the extracted query windows, we incorporate the user interaction scheme in~\cite{Huberman2016} to obtain their true labels, such that the user's visual effort and number of mouse-clicks is minimized. In this scheme, the query windows are highlighted by a green or a red frame (as shown in Figure~\ref{fig:flow}), depending on a tentative label obtained from the classifier being trained. The user is therefore only required to click inside the misclassified windows.

Due to the high degree of visual complexity that we expect, we further add a right-mouse click for signaling a ``cannot determine" user input to avoid the acquisition of incorrect labels. Note that this scheme does not require the user to inspect \emph{all} the object occurrences in the image in order to identify misclassifications, and it allows the number of mouse clicks to be significantly lower than the number of labels collected. In theory, if all the tentative labels are found to be correct, no mouse click is needed. The labels acquired in this step (excluding the rare cases of undetermined windows) are added to the training sets \posl and \negl.

Recall that the queries presented to the user are derived from \wwp and \wwn which correspond to two sets of queries, one associated with positive previously-labeled queries, and the second with negative previously-labeled queries respectively. We use this association to derive the tentative labels that we present.

As reported in the ablation study, this choice reduces the average number of user mouse clicks (corrections) by 25.4\% compared to using an all-positive query labeling. The latter would have been the case if the labels were derived from network's classification, $C(\textbf{x})$, on these windows.

\subsection{Classifier Update} At the last step of each active-learning iteration we further train the network to learn all the labels collected. This is done by performing additional ADAM minimization steps over the combined loss, $L_{total} = L_{label}+\alpha*L_{sub}$, until the labeling conditions are met, i.e., the classifications of the network are $\geq 0.95$ at the coordinates in $\mathcal{P}$, and $\leq -0.95$ at $\mathcal{N}$. Our hyper-parameters study shows that $\alpha=1$ achieves the best results. An example of the resulting network classification map, $C(\textbf{x})$, is shown in figures~\ref{fig:flow} and~\ref{fig:network}. 

\subs{Final Object Detection} Similarly to other user-assisted counting methods, such as~\cite{Arteta14}, the decision when to terminate the active-learning iterations is left to the user. As future work, we intend to explore the option of terminating the process once very few or no user corrections are needed.

The list of object occurrences outputted by our algorithm consists of all the positive non-maximum suppressed pixels in the final classification map $ C(\textbf{x})$ of the trained network.

Our method's pseudo-code can be found in the Appendix of this submission.

\section{Results}
\label{sec:results}
We implemented our algorithm in Tensorflow and evaluated it on an Nvidia GTX 1080Ti GPU. The network architecture and other algorithm parameters are all listed in Section~\ref{sec:method}. 
The set of test images used for evaluating our method consists of 47 images which we collected from a range of sources and classes. Specifically, this set includes synthetic and real fluorescence microscopy cell images from~\cite{Bernardis11,Lehmussola07}, crowd images taken from the Shanghaitech dataset~\cite{Zhang2016}, images from the Small Object dataset~\cite{Ma15}, as well as ones collected from the Web. The latter contains different types of repeating objects, such as cars, animal flocks, drink cans, food items, and people. The number of repeating instances varies considerably and ranges between 33 up to 877. Finally, the images were cropped such that all the objects are within the image boundaries, i.e., the images do not contain partial instances. These images, along with our results, can be found in the Supplemental Material of this submission. 



We report here the results of a thorough user-study that we conducted in order to evaluate our method against existing state-of-the-art user-assisted methods. Moreover, we run automated hyper-parameter search to optimize the method's performance as well as an ablation study to evaluate the contribution of the different method's components (as noted above, these tests can be found in the Appendix Section). While we do not operate in a labels-rich regime, we also compare our method against fully-supervised networks on a few images taken from their dataset. 


\subs{User-Study} We evaluated our method against the user-assisted methods of Arteta \etal~\cite{Arteta14} and Huberman and Fattal~\cite{Huberman2016} via a user-study consisting of 30 participants over 33 images. The users are novice with respect to using our method, but are accustomed to use computers on a daily basis. Before performing their task, the users received a two-minutes demonstration on how to use the interface of each method, and were instructed to perform five interactive iterations. The study was designed to produce five experiments for each method on each image in our test set. We made sure that the same user does not repeat the same experiment (image and method). Finally, the experiments were conducted in a random ordering of the methods and images, and the users were unaware as to which method is ours. To the best of our knowledge, this is the largest study reported for evaluating interactive counting methods for single-image scenarios.

\setlength{\tabcolsep}{2pt}

\begin{table*}[t]
\centering

\small
\begin{adjustbox}{width=\textwidth}
\begin{tabular}{l|ccccc|l|ccccc|l|ccccc}
\multicolumn{1}{c|}{} & \multicolumn{5}{c|}{\textbf{Arteta \etal}} &  & \multicolumn{5}{c|}{\textbf{Huberman and Fattal}} &  & \multicolumn{5}{c}{\textbf{Ours}} \\
\textbf{Image} & \textbf{\begin{tabular}[c]{@{}c@{}}Cnt. Er.\\ {[}\%{]}\end{tabular}} & \textbf{\begin{tabular}[c]{@{}c@{}}Loc. Er.\\ {[}\%{]}\end{tabular}} & \textbf{\begin{tabular}[c]{@{}c@{}}F1\\ {[}\%{]}\end{tabular}} & \textbf{\begin{tabular}[c]{@{}c@{}}Time\\ {[}sec{]}\end{tabular}} & \textbf{\begin{tabular}[c]{@{}c@{}}Clicks\\ {}{}\end{tabular}} &  & \textbf{\begin{tabular}[c]{@{}c@{}}Cnt. Er.\\ {[}\%{]}\end{tabular}} & \textbf{\begin{tabular}[c]{@{}c@{}}Loc. Er.\\ {[}\%{]}\end{tabular}} & \textbf{\begin{tabular}[c]{@{}c@{}}F1\\ {[}\%{]}\end{tabular}} &  \textbf{\begin{tabular}[c]{@{}c@{}}Time\\ {[}sec{]}\end{tabular}} & \textbf{\begin{tabular}[c]{@{}c@{}}Clicks\\ {}{}\end{tabular}} &  & \textbf{\begin{tabular}[c]{@{}c@{}}Cnt. Er.\\ {[}\%{]}\end{tabular}} & \textbf{\begin{tabular}[c]{@{}c@{}}Loc. Er.\\ {[}\%{]}\end{tabular}} & \textbf{\begin{tabular}[c]{@{}c@{}}F1\\ {[}\%{]}\end{tabular}} & \textbf{\begin{tabular}[c]{@{}c@{}}Time\\ {[}sec{]}\end{tabular}} & \textbf{\begin{tabular}[c]{@{}c@{}}Clicks\\ {}{}\end{tabular}} \\ \hline

\textbf{Antarctica} & 24.9 & 74.1 & 51.9 & \textbf{177.4} & \textbf{14.6} & & -- & -- & -- & -- & -- & & \textbf{7.3} & \textbf{27.3} & \textbf{86.6} & 210.4 & 22.0 \\
\textbf{Beach} & 7.8 & 33.3 & 77.1 & 186.5 & 20.3 &  & \textbf{4.7} & 18.2 & 90.6 & 263.8 & 25.0 &  & 5.7 & \textbf{12.2} & \textbf{94.0} & \textbf{178.0} & \textbf{18.0} \\
\textbf{Beer} & 8.7 & 22.6 & 88.1 & 174.3 & 19.7 &  & \textbf{0.5} & 5.3 & 97.3 & \textbf{53.5} & \textbf{3.0} &  & 0.9 & \textbf{1.3} & \textbf{99.4} & 96.3 & 10.4 \\
\textbf{Bees} & 19.3 & 74.3 & 54.1 & 261.0 & 16.5 &  & -- & -- & -- & -- & -- & & \textbf{8.3} & \textbf{38.0} & \textbf{80.4} & \textbf{225.7} & \textbf{16.0} \\
\textbf{Birds} & \textbf{9.2} & 23.2 & 87.0 & 206.4 & 24.8 &  & 29.0 & 47.5 & 76.0 & 290.7 & 34.0 &  & 10.9 & \textbf{19.1} & \textbf{90.4} & \textbf{205.7} & \textbf{20.8} \\
\textbf{Birds002} & 11.6 & 38.3 & 76.7 & \textbf{267.4} & 29.2 &  & 28.9 & 33.7 & 80.3 & 319.7 & 36.0 &  & \textbf{10.9} & \textbf{16.4} & \textbf{91.4} & 281.5 & \textbf{18.4} \\
\textbf{Candles} & 5.6 & 60.3 & 59.1 & 165.2 & 22.8 &  & 7.1 & 19.9 & 20.6 & \textbf{72.7} & \textbf{9.0} &  & \textbf{5.1} & \textbf{7.7} & \textbf{96.1} & 87.8 & 11.5 \\
\textbf{Cars} & 12.9 & 40.3 & 72.2 & 148.9 & \textbf{18.6} &  & 19.0 & 52.4 & 75.2 & 124.3 & 35.7 &  & \textbf{3.3} & \textbf{5.9} & \textbf{97.1} & \textbf{123.6} & 18.8 \\
\textbf{CarsBg} & 26.7 & 60.3 & 57.3 & 288.0 & 27.7 &  & \textbf{3.5} & 11.0 & 94.6 & 371.6 & 25.7 &  & 4.3 & \textbf{6.0} & \textbf{97.1} & \textbf{191.5} & \textbf{22.5} \\
\textbf{CellLrg} & 10.9 & 13.4 & 92.7 & 209.0 & 18.4 &  & 0.4 & 0.4 & 99.4 & \textbf{61.9} & 9.5 &  & \textbf{0.1} & \textbf{0.1} & \textbf{100.0} & 79.0 & \textbf{8.8} \\
\textbf{CellSml} & 3.1 & 11.9 & 94.9 & 163.0 & 19.5 &  & 0.4 & 1.3 & 99.4 & \textbf{85.2} & \textbf{6.0} &  & \textbf{0.3} & \textbf{0.3} & \textbf{99.8} & 100.8 & 9.4 \\
\textbf{Chairs} & 4.6 & 19.5 & 89.0 & 192.4 & 28.6 &  & 14.9 & 25.6 & 86.0 & 264.9 & 27.7 &  & \textbf{0.6} & \textbf{0.8} & \textbf{99.6} & \textbf{120.8} & \textbf{12.0} \\
\textbf{CokeDiet} & 8.0 & 79.5 & 50.9 & 180.0 & 12.0 &  & -- & -- & -- &--  & -- &  & \textbf{6.3} & \textbf{9.3} & \textbf{95.2} & \textbf{112.3} & \textbf{7.8} \\
\textbf{CokeReg} & 8.8 & 43.0 & 74.4 & 160.8 & 12.0 &  & -- & -- & -- & -- & -- &  & \textbf{2.4} & \textbf{2.4} & \textbf{98.9} & \textbf{77.7} & \textbf{3.4} \\
\textbf{Cookies} & 5.8 & 56.2 & 64.0 & 157.8 & 20.0 &  & 4.3 & 9.2 & 95.2 & \textbf{83.0} & 26.5 &  & \textbf{1.8} & \textbf{4.9} & \textbf{97.5} & 103.4 & \textbf{13.4} \\
\textbf{Crabs} & 4.3 & 8.7 & 94.2 & 210.3 & \textbf{14.5} &  & 27.8 & 74.7 & 62.0 & 231.9 & 37.3 &  & \textbf{2.4} & \textbf{3.0} & \textbf{98.5} & \textbf{131.8} & 15.2 \\
\textbf{Crowd} & \textbf{6.7} & 53.7 & 66.5 & 183.0 & 26.6 &  & 16.6 & 70.5 & 61.5 & 233.4 & 37.0 &  & 8.1 & \textbf{33.9} & \textbf{82.5} & \textbf{172.3} & \textbf{20.6} \\
\textbf{Discussion} & 24.5 & 112.3 & 39.3 & \textbf{187.1} & \textbf{16.9} &  & -- & -- & -- & -- & -- & & \textbf{8.9} & \textbf{53.3} & \textbf{73.0} & 262.3 & 21.2 \\
\textbf{Fish097} & \textbf{6.0} & 29.9 & 83.8 & 116.0 & 13.7 &  & 10.3 & 25.9 & 86.7 & \textbf{70.2} & \textbf{12.0} &  & 10.0 & \textbf{16.9} & \textbf{92.0} & 137.5 & 15.8 \\
\textbf{Fish107} & 14.1 & 21.0 & 88.7 & 144.5 & 16.5 &  & 10.3 & 16.2 & 92.3 & 76.5 & \textbf{8.0} &  & \textbf{2.6} & \textbf{4.8} & \textbf{97.7} & \textbf{75.6} & 9.8 \\
\textbf{Flowers} & \textbf{18.2} & 54.0 & 63.3 & 173.0 & 25.0 &  & 18.5 & 71.9 & 64.6 & 278.1 & 36.2 &  & 20.2 & \textbf{30.2} & \textbf{83.2} & \textbf{149.7} & \textbf{13.8} \\
\textbf{Hats} & 8.1 & 36.3 & 78.9 & 224.3 & 19.8 &  & 26.9 & 81.0 & 60.1 & 293.5 & 31.8 &  & \textbf{2.8} & \textbf{17.0} & \textbf{91.5} & \textbf{142.4} & \textbf{18.8} \\
\textbf{Logs} & 9.8 & 47.0 & 69.1 & 154.0 & 20.9 &  & 5.6 & 6.1 & 97.0 & 142.8 & \textbf{13.0} &  & \textbf{2.2} & \textbf{5.0} & \textbf{97.5} & \textbf{107.5} & 22.4 \\
\textbf{Matches} & 4.9 & 28.0 & 83.5 & 194.6 & 31.8 &  & 0.5 & 0.8 & 99.6 & \textbf{32.0} & 9.5 &  & \textbf{0.2} & \textbf{0.2} & \textbf{99.9} & 86.7 & \textbf{8.8} \\
\textbf{Oranges} & 11.7 & 32.8 & 82.0 & 197.6 & \textbf{19.6} &  & 20.2 & 55.8 & 69.2 & 292.6 & 26.8 &  & \textbf{5.9} & \textbf{15.7} & \textbf{92.2} & \textbf{108.9} & 20.6 \\
\textbf{Parasol} & 13.6 & 59.5 & 68.4 & 155.0 & 12.5 &  & 12.0 & 17.0 & 92.5 & 192.0 & \textbf{9.3} &  & \textbf{5.6} & \textbf{11.2} & \textbf{94.5} & \textbf{93.9} & 12.2 \\
\textbf{Peas} & 24.5 & 56.5 & 60.3 & 153.6 & 21.2 &  & 16.3 & 29.5 & 86.4 & 174.3 & 30.0 &  & \textbf{15.5} & \textbf{20.0} & \textbf{89.1} & \textbf{123.8} & \textbf{14.3} \\
\textbf{Pills} & 6.7 & 5.2 & 97.3 & 153.8 & 19.6 &  & 1.1 & 3.3 & 98.3 & \textbf{60.0} & 19.5 &  & \textbf{0.2} & \textbf{0.2} & \textbf{99.9} & 73.0 & \textbf{8.8} \\
\textbf{RealCells} & 6.0 & 32.4 & 80.5 & 139.8 & 21.8 &  & \textbf{0.6} & 5.8 & 97.1 & \textbf{95.0} & \textbf{12.5} &  & 0.8 & \textbf{3.8} & \textbf{97.5} & 175.2 & 14.4 \\
\textbf{Sheep} & 10.0 & 51.5 & 66.8 & 236.2 & 24.7 &  & 15.1 & 57.7 & 73.3 & \textbf{227.5} & 31.3 &  & \textbf{9.6} & \textbf{26.0} & \textbf{87.4} & 262.5 & \textbf{21.5} \\
\textbf{Soldiers} & 28.6 & 72.9 & 47.4 & 167.2 & 24.4 &  & 18.4 & 57.8 & 68.2 & 193.2 & 23.0 &  & \textbf{12.7} & \textbf{46.4} & \textbf{75.6} & \textbf{140.6} & \textbf{22.6} \\
\textbf{Wall} & 14.7 & 73.3 & 46.2 & 155.8 & 22.4 &  & 7.3 & 15.5 & 92.4 & \textbf{93.1} & \textbf{16.0} &  & \textbf{7.0} & \textbf{12.7} & \textbf{93.8} & 100.5 & 19.2 \\
\textbf{Water} & 10.9 & 12.6 & 93.0 & 174.4 & 17.2 &  & 1.5 & 2.3 & 98.9 & \textbf{51.9} & 11.0 &  & \textbf{0.3} & \textbf{0.3} & \textbf{99.9} & 95.8 & \textbf{6.2} \\ \hline
\textbf{Average} & 11.9 & 43.6 & 72.7 & 183.6 & 20.4 & \textbf{} & 11.5 & 29.2 & 82.7 & 168.9 & 21.5 & \textbf{} & \textbf{5.6} & \textbf{13.7} & \textbf{93.0} & \textbf{140.4} & \textbf{15.1}
\end{tabular}
\end{adjustbox}

\normalfont
\vspace{3mm}

\caption{User-Study Results. The columns report (left-to-right): the average counting error percentage, localization error percentage (false-positives plus false-negatives), F-score, interactive session time and the number of user mouse-clicks. This is repeated for each method. The most accurate counting and localization estimates are marked in bold. The rows list the images used. Images Birds002, Fish097, Fish107, and Bees were taken from the Small Objects dataset~\cite{Ma15}, CellLrg and CellSml from~\cite{Lehmussola07}, Soldiers from~\cite{Zhang2016} and RealCells from~\cite{Bernardis11}.}
\label{tab:comp}

\end{table*}

Table~\ref{tab:comp} presents the results obtained by this study, and includes the counting error, two measures of localization error, the number of user mouse-clicks and the overall session time for each method and test image. Note that F-score measures accuracy hence higher is better. A more comprehensive table is provided in the Appendix Section. In order to evaluate the localization, we carefully prepared ground-truth maps where each object appearance is recorded at its center of its mass. The predicted location of an object is considered to be correct if its surrounding 21-by-21 pixels window contains an entry in the ground-truth map. 

\begin{figure}[t]
\centering
\includegraphics[width=3.5in]{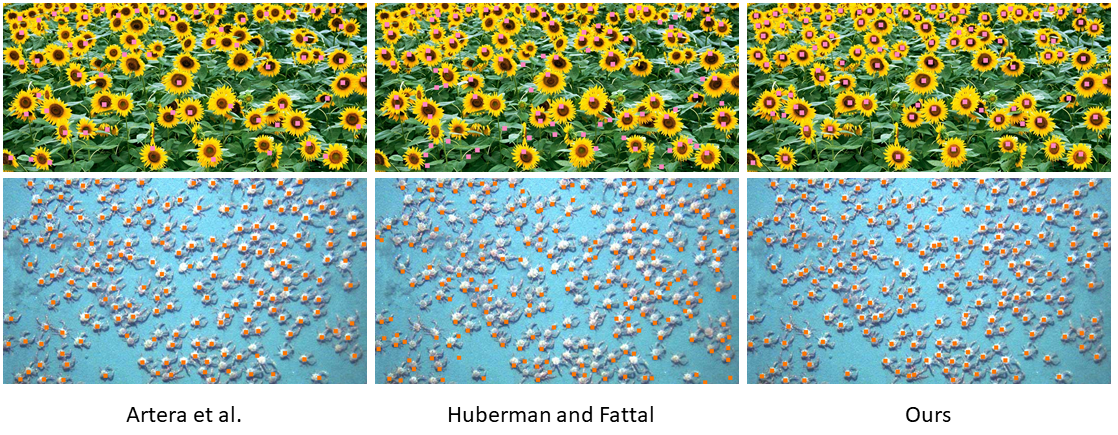}
\caption{Localization Example. The results of Arteta \etal~\cite{Arteta14}, Huberman and Fattal~\cite{Huberman2016}, and our methods are shown on the Flowers and Crabs images.}
\label{fig:results}
\end{figure}

On average, the counting error percentage indicates that our method achieves a greater counting accuracy compared to the methods of Arteta \etal~\cite{Arteta14} and Huberman and Fattal~\cite{Huberman2016}, and stands on 5.6\%, 11.9\% and 11.5\% respectively. Huberman and Fattal's method appears to fail on cases where the object of interest, due to its high complexity, does not consist of many repeating patches or they do not show sufficient spatial correlation. These images are omitted from Table~\ref{tab:comp}. The standard deviation of the error percentages in both Arteta \etal's and our methods (as shown in the full table), which stand on 7.1\% and 4.8\% respectively, imply a fair amount of consistency by these methods. 

We emphasize that in these tests our method was trained and tested over a single image, assuming no prior knowledge about the object type. The set of images used includes a wide range of image categories that demonstrate the method's flexibility and applicability.

Our improved counting accuracy stems from a considerable reduction in the localization error. When summing the false-positives and false-negatives predictions, our methods stands on 13.7\% average error compared to 43.6\% by Arteta el al.'s method, and 29.2\% by Huberman and Fattal's. As shown in the more comprehensive comparison table, Arteta el al.'s average error precision is fairly high (89.3\%) but their recall is considerably lower (63.5\%), which suggests that their method misses many object appearances. This large error is inconsistent with the lower false-negative error implied by their counting error. Indeed, this method performs its counting and localization estimates independently. Figure~\ref{fig:results} shows example object localizations produced by the three methods. Our output appears to be well-correlated with the object's appearances. 

In terms of average session times, our method compares favorably, where the results discussed above were achieved by our method in an average session time of 140 seconds, by Arteta \etal's method 183 seconds, and in Huberman and Fattal's 168 seconds. These session times reported include the initialization time, interaction time of five iterations as well as training time of each iteration. It should be noted that our method uses a richer visual model, but this is compensated by its GPU implementation as well as the user-friendly interface borrowed from~\cite{Huberman2016} for collecting the user feedback. These improved session time also correlate with the number of recorded mouse-clicks, where the users performed 15.1 clicks on average when using our method, 20.4 clicks when using Arteta \etal, and 21.5 when using Huberman and Fattal's.

Some of the images belong to the same source (scene) and allow us to evaluate the performance of a model, trained on one image, over the rest of the images. Specifically, train models over the CellSml, Birds002 and Bees (from the user-study above) and test them over different 7 cells images, 2 birds images and 5 bees images respectively. The resulting average counting error and F-score are: 0.4 and 99.7 on CellSml images, 11 and 91.5 over the Birds, and 7.74 and 84.3 for the Bees. These results show our method generalizes well over images from the same source, and provides an attractive solution for such a scenario.

Finally, let us note that CokeReg and CokeDiet in Table~\ref{tab:comp} correspond to the same image containing two types of cans. Based on the user's input bounding-windows and labelings, our network successfully counts one type or the other.

While the method of von Borstel \etal~\cite{Borstel16} does not provide object localization, it relies on a fairly small number of labels to perform object counting. On the fluorescence microscopy cell dataset~\cite{Lempitsky2010}, von Borstel \etal report a counting average error of 6.7 cells given 70 user-provided labels, where our method resulted in an average error of 6.06 cells, or 4.6\% average counting error percentage, using the same number of labels. Moreover, our localization error percentage and F-score average are 13.2\% and 93.2\% respectively, which is consistent with the errors reported in Table~\ref{tab:comp}. These values are unavailable for the method of von Borstel \etal as it does not localize the objects.
 
\subs{Supervised Methods} While our method operates in a different training setting and targets a different practical scenario, we provide a partial comparison to several fully-supervised counting networks trained on large datasets. 

The method in~\cite{Zhang2016} estimates the number of people in crowd scenes and was trained over a few hundreds of images. We included in our dataset one such image, titled Crowd, on which we applied this method. The counting error produced by this network is 14.1\%, while our weakly-supervised approach results in 8.1\% on average. We believe this supports the claim that the extra effort in running an interactive approach pays off when the counting is required only over a small number of images of a particular class.

The network in~\cite{Ma15} is used for counting and localizing small objects, and was trained independently over four different classes (fish, bees, flies and seagull). Our dataset contains some images from this dataset, namely, Bees, Fish097, Fish107 and Birds002. When training over about 300 training labels, Ma \etal~\cite{Ma15} obtain an average F-measure of 84.1\% over the bees images, and an average of 87.7\% over the fish images. Our method achieves averages of 80\% and 93\% on these image types respectively. In the case of the seagull class (Birds002 in our set), they use about 930 labels and obtain an F-measure average of 88.6\%, where our average is 91\%. The average number of mouse-clicks provided by our users is 15. 


In order to avoid training a classifier for specific object classes, Lu \etal~\cite{Lu2018} train a generic matching network, combine it with an adapter module that specializes on the input class given a small number of labels. They report an average counting error of $3.56\pm0.27$ over cells images containing between 74 and 317 occurrences, implying an average percentage error of at least $0.78\%$ to $3.35\%$ which is $\times2.6$ to $\times11$ higher than ours.



\section{Conclusions}
We presented a new weakly-supervised CNN training for localizing and counting repeating objects applicable for single image scenarios. This is made practical using an iterative active-learning procedure that minimizes the number of labels collected by analyzing the latent representations of the network in order to avoid querying closely-related windows and ones which are already labeled.

Evaluation against existing user-assisted counting methods demonstrates the network ability to achieve state-of-the-art performance both in terms of counting and localization accuracy. We conducted a user-study that demonstrated the method's superior performance on a wide range of image classes and visual levels of complexity.
 


{\small
\bibliographystyle{ieee_fullname}
\bibliography{egbib}

\begin{thebibliography}{10}\itemsep=-1pt

\bibitem{Arteta14}
Carlos Arteta, Victor~S. Lempitsky, J.~Alison Noble, and Andrew Zisserman.
\newblock Interactive object counting.
\newblock In {\em Proceedings of the European Conference on Computer Vision
  (ECCV)}, pages 504--518, 2014.

\bibitem{Bernardis11}
Elena Bernardis and Stella Yu.
\newblock Pop out many small structures from a very large microscopic image.
\newblock {\em Medical image analysis}, 15:690--707, 2011.

\bibitem{Borstel16}
M.~V. Borstel, M. Kandemir, Philip Schmidt, Madhavi~K. Rao, K. Rajamani, and F.
  Hamprecht.
\newblock Gaussian process density counting from weak supervision.
\newblock In {\em Proceedings of the European Conference on Computer Vision
  (ECCV)}, pages 365--380, 2016.

\bibitem{Cao2017}
Liujuan Cao, Feng Luo, Li Chen, Yihan Sheng, Haibin Wang, Cheng Wang, and
  Rongrong Ji.
\newblock Weakly supervised vehicle detection in satellite images via
  multi-instance discriminative learning.
\newblock {\em International Conference on Pattern Recognition (ICPR)},
  64:417–424, 2017.

\bibitem{Caron18}
Mathilde Caron, Piotr Bojanowski, Armand Joulin, and Matthijs Douze.
\newblock Deep clustering for unsupervised learning of visual features.
\newblock In {\em Proceedings of the European Conference on Computer Vision
  (ECCV)}, volume 11218, pages 139--156, 2018.

\bibitem{Chan08}
{Antoni B.} Chan, {Zhang-Sheng John} Liang, and Nuno Vasconcelos.
\newblock Privacy preserving crowd monitoring: Counting people without people
  models or tracking.
\newblock In {\em Proceedings of the IEEE Conference on Computer Vision and
  Pattern Recognition (CVPR)}, 2008.

\bibitem{Chen2012}
Ke Chen, Chen~Change Loy, Shaogang Gong, and Tony Xiang.
\newblock Feature mining for localised crowd counting.
\newblock In {\em The British Machine Vision Conference (BMVC)}, pages
  21.1--21.11, 2012.

\bibitem{Cohen2017}
Joseph~Paul Cohen, G. Boucher, Craig~A. Glastonbury, Henry~Z. Lo, and Yoshua
  Bengio.
\newblock Count-ception: Counting by fully convolutional redundant counting.
\newblock {\em Proceedings of the IEEE International Conference on Computer
  Vision (ICCV) Workshops}, 2017.

\bibitem{Dijkstra18}
Klaas Dijkstra, Jaap van~de Loosdrecht, L.~R.~B. Schomaker, and Marco~A.
  Wiering.
\newblock Centroidnet: {A} deep neural network for joint object localization
  and counting.
\newblock In {\em European Conference on Machine Learning and Principles and
  Practice of Knowledge Discovery in Databases (ECML-PKDD)}, pages 585--601,
  2018.

\bibitem{Dong07}
Lan Dong, Vasu Parameswaran, Visvanathan Ramesh, and Imad Zoghlami.
\newblock Fast crowd segmentation using shape indexing.
\newblock {\em Proceedings of the IEEE International Conference on Computer
  Vision (ICCV)}, 2007.

\bibitem{Girshick2014}
Ross Girshick, Jeff Donahue, Trevor Darrell, and Jitendra Malik.
\newblock Rich feature hierarchies for accurate object detection and semantic
  segmentation.
\newblock In {\em Proceedings of the IEEE Conference on Computer Vision and
  Pattern Recognition (CVPR)}, pages 580--587, 2014.

\bibitem{Hsieh2017}
Meng-Ru Hsieh, Yen-Liang Lin, and Winston~H. Hsu.
\newblock Drone-based object counting by spatially regularized regional
  proposal networks.
\newblock In {\em Proceedings of the IEEE International Conference on Computer
  Vision (ICCV)}, pages 4145--4153, 2017.

\bibitem{Hu18}
J. Hu, L. Shen, and G. Sun.
\newblock Squeeze-and-excitation networks.
\newblock In {\em Proceedings of the IEEE Conference on Computer Vision and
  Pattern Recognition (CVPR)}, pages 7132--7141, 2018.

\bibitem{Huberman2016}
Inbar Huberman and Raanan Fattal.
\newblock Detecting repeating objects using patch correlation analysis.
\newblock In {\em Proceedings of the IEEE Conference on Computer Vision and
  Pattern Recognition (CVPR)}, volume~34, 2016.

\bibitem{Idrees2013}
Haroon Idrees, Imran Saleemi, Cody Seibert, and Mubarak Shah.
\newblock Multi-source multi-scale counting in extremely dense crowd images.
\newblock In {\em Proceedings of the IEEE Conference on Computer Vision and
  Pattern Recognition (CVPR)}, 2013.

\bibitem{Johnson16}
Justin Johnson, Alexandre Alahi, and Li Fei-Fei.
\newblock Perceptual losses for real-time style transfer and super-resolution.
\newblock In {\em Proceedings of the European Conference on Computer Vision
  (ECCV)}, 2016.

\bibitem{Kang17}
Di Kang, Zheng Ma, and Antoni Chan.
\newblock Beyond counting: Comparisons of density maps for crowd analysis tasks
  - counting, detection, and tracking.
\newblock {\em IEEE Transactions on Circuits and Systems for Video Technology
  (TCSVT)}, 2017.

\bibitem{Kong06}
Dan Kong, Doug Gray, and Hai Tao.
\newblock A viewpoint invariant approach for crowd counting.
\newblock In {\em International Conference on Pattern Recognition (ICPR)},
  volume~3, 2006.

\bibitem{Krizhevsky2012}
Alex Krizhevsky, Ilya Sutskever, and Geoffrey~E. Hinton.
\newblock Imagenet classification with deep convolutional neural networks.
\newblock In {\em Advances in Neural Information Processing Systems (NeurIPS)},
  pages 1097--1105, 2012.

\bibitem{Kupyn2018}
O. {Kupyn}, V. {Budzan}, M. {Mykhailych}, D. {Mishkin}, and J. {Matas}.
\newblock Deblurgan: Blind motion deblurring using conditional adversarial
  networks.
\newblock In {\em Proceedings of the IEEE Conference on Computer Vision and
  Pattern Recognition (CVPR)}, pages 8183--8192, 2018.

\bibitem{Lehmussola07}
Antti Lehmussola, Pekka Ruusuvuori, Jyrki Selinummi, Heikki Huttunen, and Olli
  Yli-Harja.
\newblock Computational framework for simulating fluorescence microscope images
  with cell populations.
\newblock {\em IEEE Transactions on Medical Imaging}, 26(7):1010--1016, 2007.

\bibitem{Lempitsky2010}
Victor Lempitsky and Andrew Zisserman.
\newblock Learning to count objects in images.
\newblock In {\em Advances in Neural Information Processing Systems (NeurIPS)},
  volume~23, pages 1324--1332, 2010.

\bibitem{Loy13}
Chen~Change Loy, Shaogang Gong, and Tao Xiang.
\newblock From semi-supervised to transfer counting of crowds.
\newblock In {\em Proceedings of the IEEE International Conference on Computer
  Vision (ICCV)}, pages 2256--2263, 2013.

\bibitem{Lu2018}
E. Lu, W. Xie, and A. Zisserman.
\newblock Class-agnostic counting.
\newblock In {\em Asian Conference on Computer Vision (ACCV)}, 2018.

\bibitem{Ma15}
Z. {Ma}, {Lei Yu}, and A.~B. {Chan}.
\newblock Small instance detection by integer programming on object density
  maps.
\newblock In {\em Proceedings of the IEEE Conference on Computer Vision and
  Pattern Recognition (CVPR)}, pages 3689--3697, 2015.

\bibitem{quteprints09}
David Ryan, Simon Denman, Clinton~B. Fookes, and Sridha Sridharan.
\newblock Crowd counting using multiple local features.
\newblock In {\em Digital Image Computing : Techniques and Applications
  (DICTA)}, 2009.

\bibitem{Scheffer01}
Tobias Scheffer, Christian Decomain, and Stefan Wrobel.
\newblock Active hidden markov models for information extraction.
\newblock In {\em Proceedings of International Conference on Advances in
  Intelligent Data Analysis}, pages 309--318, 2001.

\bibitem{SeguiPV15}
S. {Seguí}, O. {Pujol}, and J. {Vitrià}.
\newblock Learning to count with deep object features.
\newblock In {\em Proceedings of the IEEE Conference on Computer Vision and
  Pattern Recognition (CVPR) Workshops}, pages 90--96, 2015.

\bibitem{Walach16}
Elad Walach and Lior Wolf.
\newblock Learning to count with cnn boosting.
\newblock In {\em Proceedings of the European Conference on Computer Vision
  (ECCV)}, 2016.

\bibitem{wang2018}
Ting-Chun Wang, Ming-Yu Liu, Jun-Yan Zhu, Guilin Liu, Andrew Tao, Jan Kautz,
  and Bryan Catanzaro.
\newblock Video-to-video synthesis.
\newblock In {\em Advances in Neural Information Processing Systems (NeurIPS)},
  2018.

\bibitem{Xie16}
W. Xie, J.~A. Noble, and A. Zisserman.
\newblock Microscopy cell counting and detection with fully convolutional
  regression networks.
\newblock {\em Computer Methods in Biomechanics and Biomedical Engineering:
  Imaging \& Visualization}, 6:283--292, 2016.

\bibitem{Zhang18}
Richard Zhang, Phillip Isola, Alexei Efros, Eli Shechtman, and Oliver Wang.
\newblock The unreasonable effectiveness of deep features as a perceptual
  metric.
\newblock In {\em Proceedings of the IEEE Conference on Computer Vision and
  Pattern Recognition (CVPR)}, pages 586--595, 2018.

\bibitem{Zhang2016}
Yingying Zhang, Desen Zhou, Siqin Chen, Shenghua Gao, and Yi Ma.
\newblock Single-image crowd counting via multi-column convolutional neural
  network.
\newblock {\em Proceedings of the IEEE Conference on Computer Vision and
  Pattern Recognition (CVPR)}, pages 589--597, 2016.

\end{thebibliography}
}

\end{document}